\documentclass[conference]{IEEEtran}
\IEEEoverridecommandlockouts
\usepackage{cite}
\usepackage{amsmath,amssymb,amsfonts}
\usepackage{float}
\usepackage{booktabs}
\usepackage{multirow}
\usepackage{algorithmic}
\usepackage{graphicx}
\usepackage{textcomp}
\usepackage[numbers]{natbib}
\usepackage{subcaption}
\usepackage{xcolor}
\def\BibTeX{{\rm B\kern-.05em{\sc i\kern-.025em b}\kern-.08em
    T\kern-.1667em\lower.7ex\hbox{E}\kern-.125emX}}
    
\usepackage{tabularx}
\begin{document}

\title{Sound Event Classification in an Industrial Environment: Pipe Leakage Detection Use Case}

\author{\IEEEauthorblockN{Ibrahim Shaer}
\IEEEauthorblockA{Department of Electrical and Computer Engineering \\
\textit{Western University}\\
London, Canada \\
ishaer@uwo.ca}
\and
\IEEEauthorblockN{Abdallah Shami}
\IEEEauthorblockA{Department of Electrical and Computer Engineering \\
\textit{Western University}\\
London, Canada \\
abdallah.shami@uwo.ca}
}

\maketitle

\begin{abstract}
In this work, a multi-stage Machine Learning (ML) pipeline is proposed for pipe leakage detection in an industrial environment. As opposed to other industrial and urban environments, the environment under study includes many interfering background noises, complicating the identification of leaks. Furthermore, the harsh environmental conditions limit the amount of data collected and impose the use of low-complexity algorithms. To address the environment's constraints, the developed ML pipeline applies multiple steps, each addressing the environment's challenges. The proposed ML pipeline first reduces the data dimensionality by feature selection techniques and then incorporates time correlations by extracting time-based features. The resultant features are fed to a Support Vector Machine (SVM) of low-complexity that generalizes well to a small amount of data. An extensive experimental procedure was carried out on two datasets, one with background industrial noise and one without, to evaluate the validity of the proposed pipeline. The SVM hyper-parameters and parameters specific to the pipeline steps were tuned as part of the experimental procedure. The best models obtained from the dataset with industrial noise and leaks were applied to datasets without noise and with and without leaks to test their generalizability. The results show that the model produces excellent results with 99\% accuracy and an F1-score of 0.93 and 0.9 for the respective datasets.

\end{abstract}

\begin{IEEEkeywords}
Industrial Internet of Things, Sound Event Detection, Industrial Event Detection, Machine Learning Pipeline, Support Vector Machine
\end{IEEEkeywords}

\section{Introduction}
The Internet of Things (IoT) technologies have been widely adopted in economic, industrial, and health sectors given by their promise of sensing, automating, and actuating processes \cite{b21}. As part of their perceived function, IoT sensors can classify gathered data to rationalize the phenomena monitored over an extended period. The task of classifying time-series data gives rise to Time Series Classification (TSC) as a challenging aspect of the data mining field. In that regard, the classification of the gathered data can detect the existence or absence of the monitored phenomena. 

Audio Event Classification (AEC) represents one application leveraging the data collected by IoT devices. Sound classification applications span automatic speech recognition \cite{b27}, and Environment Sounds Classification (ESC) \cite{b15}. In comparison to other disciplines, non-stationarity and aperiodicity characterise ESC tasks \cite{b28}. The events in an urban or industrial environment, belonging to the class of ESC tasks, display no specific structure and are randomly generated as opposed to music and speech signals. Such factors challenge identifying sound events of interest in an industrial environment. 

In an industrial facility, stand-alone or fluid-carrying pipes integrated into larger systems are essential for manufacturing processes \cite{survey_leakage}. The processes result in unfavourable conditions for the facility's workers such as the exposure to high temperatures or the inhalation of harmful material, often rendering the area around these pipes unreachable for humans. These conditions hinder the workers' ability to identify any leaks by eye test or manual experimentation. As such, augmenting pipes with sensors to collect nearby acoustic variations and linking them to the leakage phenomenon can be achieved using Industrial IoT (IIoT) devices. To extract the concealed relationships between the collected acoustic data and the presence of leaks, supervised Machine Learning (ML) techniques can be utilized.  

The leakage detection problem has been previously addressed in literature in the works of Beghi \textit{et al.} \cite{b29} and in real-world implementations of the Wessex county water leakage detection challenge \cite{b30}. In comparison with these environments, the industrial setup under study exhibits two distinctive characteristics. First, the industrial framework encompasses many interfering sounds that can potentially mask the identification of leaks. Such noises may include the manufacturing process,  which can be broken down into multiple phases, each displaying its unique characteristics. Second, the conditions created by the manufacturing process challenge extensively running experiments that simulate leak conditions; thus, contributing to a limited amount of collected data. Moreover, the developed module should be deployed on gateways of limited computational capabilities close to the sensors collecting the data. This aspect is prevalent in the edge computing paradigm \cite{shaer2020multi} that is adopted in the realm of IIoT applications to mitigate the communication issues between sensors and gateways \cite{khan2020industrial}.

The data pertaining to our study were collected in an industrial setup in collaboration with an industry partner, and the fluid leakages were manually created. The acoustic data were represented in the time-frequency domain to address the high-dimensionality of the raw acoustic signals. To address the challenges imposed by the industrial setting under study and its requirements, the contributions of this work are as follows: 
\begin{itemize}
    \item Identify the challenges associated with fluid leakage detection in an industrial setting and highlight its uniqueness in comparison to other settings;
    \item Propose an ML pipeline that addresses each of the challenges imposed by the industrial environment;
    \item Apply the proposed pipeline to three datasets, collected in different industrial settings, and explain its capacity to identify fluid leakages while minimizing false alarms.
\end{itemize}

The rest of the paper is structured as follows: Section 2 outlines the related work. Section 3 describes the methodology. Section 4 provides the experimental setup. Section 5 explains the results and Section 6 concludes the paper. 

\section{Related Work}
The problem of identifying leaks is a synergy between the TSC and ESC tasks. Therefore, this section will investigate the contributions of the research community in both disciplines. 

Research conducted to address TSC tasks have experimented with hand-crafted and automatic feature extraction techniques. The works in \cite{b6, b9, b34, xiao2021rtfn} applied Deep Neural Networks (DNNs) to time-series datasets of the UCR repository, used as a benchmark to compare different approaches. These approaches either use the raw time-series data as inputs, as is the case for the work in \cite{b6}, or transform the time-series data into images, that is adopted in the remaining works of this category. In the realm of hand-crafted feature extraction, the works in \cite{b40, b41} are the most prominent. Salamon \textit{et al.} \cite{b40} created a codebook from the time-frequency domain data to encode different frames of the data. The encoded frames based on this codebook are fed to a decision tree algorithm to classify the TSC tasks in the UCR repository. On the same dataset, the work in \cite{b41} showed that the classification accuracy of Fully Convolutional Network (FCN), when fed with statistical features, outperforms the approach that only uses the raw input data. 

The surveyed ESC tasks include works classifying polyphonic and monophonic sounds, which resemble our collected data that include overlapping leak and industrial noise instances. The approaches proposed were applied to the Detection and Classification of Acoustic Scenes and Events (DCASE) challenge, which represents the largest collection of monophonic sounds. Koutini \textit{et al.} \cite{koutini2021receptive} and  Yuji \textit{el al.} \cite{b33} are two works that applied DNNs to extract features and classify monophonic sounds. On the other hand, the authors in \cite{b12, b13} have also used DNNs to classify polyphonic sound events derived from the DCASE datasets. They synthetically created polyphonic sounds by overlapping different sounds and employing data augmentation techniques to address data scarcity issues. All of the mentioned works, except for the case of polyphonic sounds, have reported high accuracy results relative to each of their tasks, in the range of 90\%.

Evaluating the outlined approaches will target the datasets the researchers used to test their methods as they reflect the environment under study, and the algorithms used for ESC and TSC tasks. The methods that apply DNNs for both TSC and ESC tasks on either image transformations of raw input data or the input data itself essentially violate two of the main constraints of this industrial setup. First, the limited capability of gateways hinders the training and testing of DNN models due to their resource-intensive nature. Second, to avoid over-fitting, such models require large amounts of data, which are not available in the current environment. On the other hand, the datasets used to evaluate different methods do not reflect the industrial environment because there is no transition phase from an in-existent event to the event itself. To address these limitations, this study offers a light-weight ML approach that combines feature selection and feature engineering techniques applied to time-frequency data. The solution addresses the concerns related to any industrial environment. 

\section{Methodology}
This section explains the method undertaken to tackle the leakage detection problem. The first subsection describes the datasets. The second subsection details the different steps of the applied Machine Learning Pipeline. 
\subsection{Data Description}
Each of the collected datasets describes the Power Spectral Density values (PSD) of the process under study. In each dataset, the columns represent the time domain, and the rows represent the frequency domain. The proposed ML pipeline was applied to three different datasets, each with its specific length and ambient conditions. The details of each dataset are provided in Table \ref{Datasets' Details}. Separated by a dash (\_), the naming convention identifies the existence of leaks and industrial noise. The datasets were collected with a one-second granularity, and the frequencies obtained are in the range of 0 to 65,536 Hz with a sampling frequency of 131,072 Hz. The resulting datasets are of row size 5,000 and column size equivalent to the period of data collection. The last column in the table represents the leak intervals in each of the datasets. Based on Table \ref{Datasets' Details}, the Leak\_process dataset has 4 leaks, and the Leak\_noprocess dataset has 3 leaks.

\begin{table}[htbp]
  \centering
  \scriptsize
    \begin{tabularx}{\linewidth}{|c|c|X|}
    \hline
      \textbf{Name} & \textbf{Duration (sec)} & \textbf{Leaks} \\
      
      \hline  
       Leak\_process & 3096 & 
       $\left[1191:1276\right], \left[1370:1450\right],$ \newline $\left[1796:1886\right], \left[1990:2081\right]$ \\
       \hline 
       Leak\_noprocess & 3069 & 
       $\left[2300:2348\right], \left[2623:2670\right],$ \newline $\left[2783:2833\right]$ \\
       \hline 
       NoLeak\_noprocess & 2634 & N/A \\
       \hline
    \end{tabularx}
    \caption{Datasets' Details}
    \label{Datasets' Details}
\end{table}

\subsection{Machine Learning Pipeline}
The proposed ML Pipeline is depicted in Figure \ref{ML_pipeline}. Five stages of pre-processing, training, validation, and testing are applied to the dataset with leaks and manufacturing noise denoted by Leak\_process. This dataset is elected for this procedure because it encompasses a more profound environment than the other two available datasets. The other two datasets go through the same procedure, except that the model selection phase is replaced by the best model trained on the Leak\_process dataset. The description of each stage of the ML pipeline is as follows:
\begin{figure} [htbp]
    \centering
    \includegraphics[scale = 0.42]{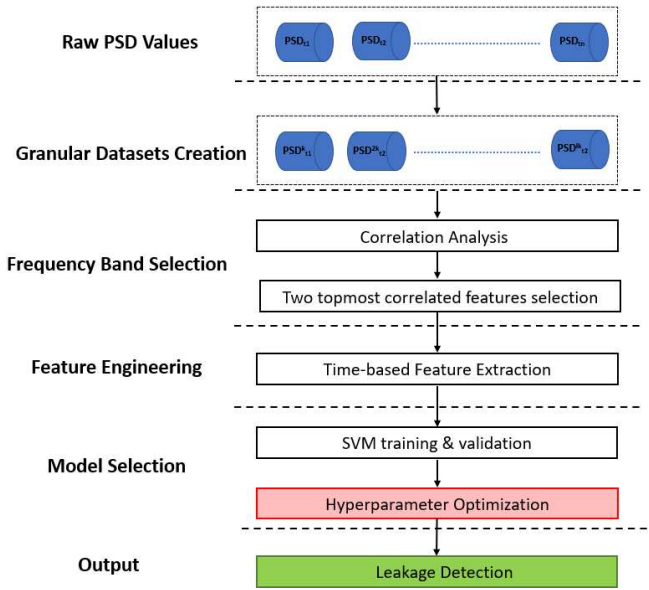}
    \caption{Machine Learning Pipeline}
    \label{ML_pipeline}
\end{figure}
\subsubsection{Granular Dataset Creation}
Each of the datasets in table \ref{Datasets' Details} is an input to this stage. The first stage creates three datasets of 1000 Hz, 2000 Hz, and 5000 Hz granularities to reduce the dimensionality of the datasets by grouping sub-bands. The value of each sub-band is calculated using three metrics: mean, median, and inter-quartile range (iqr). The metrics are chosen to study the potential effect of outliers in each sub-band on leakage detection.  

\subsubsection{Frequency Band Selection}
This stage takes as an input the granular datasets created in the previous stage. The data pre-processing necessitates the extraction of sub-bands that best represent the leakages, regardless of other interfering factors. While these factors can affect a host of sub-bands, selecting specific bands can minimize their potential effect on masking the identification of leaks. Pearson correlation is calculated to quantify the assumed linear relationship between the sub-bands and the probability of leaks. This step aligns with the goal of reducing the dimensionality of the data to accelerate the ML modelling \cite{manias2021concept}. The resulting datasets following this stage include the features with the top-most correlation values with the output variable.

\subsubsection{Feature Engineering}
This stage takes the resulting datasets from the previous step and applies a time-based feature engineering technique to each of the sub-bands. The temporal aspects are captured by overlapping adjacent time windows and calculating statistical and temporal features. 

The time-based features are listed in Table \ref{Time-based Features}. In this table, $x_i$ denotes the value of the sub-band at time step $i$ after implementing the sliding window process and $N$ represents the ten-second time frame. For $A_t$, five different values of $t$ were considered to calculate the auto-correlations: $t = \{1, 2, 3, 4, 5\}$. For $pct_t$, three different values of $t$ were considered: $t = \{1, 2, 3\}$. Determining the bins that different values of $x_i$ belong is a prerequisite for calculating the Shannon Entropy. To mirror the leakage uncertainty, the $5^{th}$, $10^{th}$, $95^{th}$, $99^{th}$ quantiles are calculated for the leak data. Accordingly, this data displays a distinctively higher entropy values compared to the data with no leaks. 

In addition to the listed features, Approximate Entropy $APEn$ and Sample Entropy $SampEn$ were included in the feature engineering procedure. Due to space constraints, the keen readers can refer to the work in \cite{b20} to check their calculation procedure. $APEn$ \cite{b20} quantifies the amount of unpredictability of fluctuations in time series data to highlight the transition phase between the two stages of absence and presence of leaks and vice versa. $SampEn$ \cite{b20} addresses the limitations pertaining to the calculation of $APEn$ by eliminating the re-calculation of a template vector, which adds a bias to the $APEn$ values. It serves a similar purpose to $APEn$ in terms of quantifying the underlying data periodicities. Due to space limitations, the full explanation of some features will be omitted. The work of Sohaib \textit{et al.} \cite{sohaib2019leakage} provides a detailed explanation of some of the adopted features.

Since the time-based features require a specific time window to be calculated, a ten-second time window is considered. A leak is identified when a period of ten seconds has passed with leak instances, reflecting that no anomalous non-leak sound events are being recorded in that time window. The resulting datasets include 52 predictors for two sub-bands, each representing a time-based feature, and one output variable representing the existence of leaks. The number of rows varies depending on the overlapping time window and the original size of the dataset. 

\begin{table*}[htbp]
    \centering
    \makebox[\linewidth]{
    \begin{tabular}{|c|c|c|c|c|c|}
    \hline \textbf{Feature Name} & \textbf{Equation} & \textbf{Feature Name} & \textbf{Equation} & \textbf{Feature Name} & \textbf{Equation}\\
    \hline
    Peak & $P=max (|x_i|)$ & Impulse Factor & $\frac{P}{\frac{1}{N} \sum_{i=1}^{N}|x_i|}$ & Square Root Mean & $SRM=(\frac{1}{N}\sum_{i=1}^{N}\sqrt{|x_i|})^2$\\[3ex]
    
    Clearance Factor & $\frac{P}{SRM}$ & 
    Root Mean Square &  $\sqrt{(\frac{1}{N} E)}$  & Margin Factor & $\frac{P}{SRM}$\\[3ex]
     Energy & $E=\sum_{i=1}^{N} x^2$ & Crest Factor & $\frac{P}{RMS} $  & Peak-to-peak & $max(x_i) - min(x_i)$ \\[3ex]
    Kurtosis & $\frac{1}{N}\sum_{i=1}^{N}(\frac{x_i-\overline{x}}{\sigma})^4$ & Skewness & $\frac{1}{N} \sum_{i=1}^{N} (\frac{x_i-\overline{x}}{\sigma}) ^ 3$ & Shape Factor &  $\frac{RMS}{\frac{1}{N} \sum_{i=1}^N |x_i|}$\\[3ex]
    
    Index Maximum & $\frac{P}{N}$ & Index Minimum &  $\frac{min (|x_i|)}{N}$ & Auto-correlation & $A_t=\frac{\sum_{i=1}^{N-t} (X_i - \bar X) (X_{i+t} - \bar X)}{\sum_{i=1}^{N} (Y_i - \bar Y) ^ 2}$\\[3ex]

    Percentage Change & $pct_t=\frac{(X_i - X_{i-t}) \times 100 }{|X_{i-t}|}$ & Shannon Entropy & $-\sum_{i=1}^{N} P(x_i) \log P(x_i)$ & Rate Entropy & $H(X_n | X_{n-1}, X_{n-2})$\\[3ex]
    
    \hline
\end{tabular}}
    \caption{Time-based Features}
    \label{Time-based Features}
\end{table*}

\subsubsection{Model Selection}
The time-based features extracted in the previous step are the input to this \textit{Model Selection} stage. This stage conducts extensive experimentation to decide the best parameters, which encompass finding the best hyper-parameters for the chosen algorithm, the size of the overlapping time window, the granularity and sub-band coupling, and the best metric to detect leaks. 

The Support Vector Classifier (SVC) algorithm is chosen to classify the data to leak and no leak instances. This algorithm is selected for multiple reasons. It is a low-complexity ML algorithm that uses linear, polynomial, and radial kernel functions. These functions map the initial feature space into a higher-dimensional space, which enables evaluating feature interactions that may contribute to improving the distinction between leak and no leak instances. Accordingly, this ``kernel trick" refines the construction of a hyper-plane with maximal margins to provide better generalization for linearly or non-linearly separable kernel functions \cite{b44}. 

The light-weight nature of the SVC algorithm favours its deployment on IIoT gateways of limited computational resources; thus, facilitating the model's re-training and testing on new data. Moreover, the scarcity of data in such a harsh environment requires a model that can generalise well with a small amount of data, which is a feature of the SVC algorithm. In the field of light-weight algorithms, Federated Learning approaches are promising in IIoT applications \cite{abdulrahman2020survey}, which will be explored in future work. The SVC algorithm determines the hyper-plane separating the involved classes using a limited amount of data. In particular, the instances that are close to the hyper-plane drive the decision boundary, which separates the two classes of the leak and no leak instances. As such, if the data points fed to the SVC model are representative of both classes, the SVC would be able to generalise to different scenarios. In this industrial setup, it is critical to provide the reasons for a particular classification result. To that end, the developed SVC model provides better interpretability for any resulting classification when compared to deep learning models. 
 
\section{Experimental Setup}
This section explains the experimental setup. It starts with outlining the evaluation criteria and explaining the parameters used for different types of datasets. Next, it discusses the experimental procedure in terms of the used parameters and selected models in light of the proposed ML pipeline. 
\subsection{Evaluation Criteria}
The models built using different parameters are evaluated using the standard performance metrics specific to a classification task. These metrics include accuracy, specificity, recall, and precision. Different metrics are prioritized depending on the testing dataset .For the Leak\_noprocess and Leak\_process datasets, precision, recall, and specificity will be used as an indicator of the models' performance. These indicators reflect the model's ability to correctly identify leak instances while minimizing false alarms. For the Leak\_noprocess dataset, a higher priority will be given to precision and recall as indicators of the quality of the model and the chosen parameters. Such considerations will be highlighted while explaining the obtained results.  

\subsection{Experimental Procedure}
The different parameters involved in the experimental procedure are summarized in Table \ref{Experimental Parameters}. As is the case in any ML modelling, the dataset is split into training, validation, and testing sets. The training set is used to train the model, which is evaluated on the validation set to select the best parameters outlined in table \ref{Experimental Parameters}. Since 864 parameters are being tuned, the hyper-parameters of the SVC model and the overlapping time window parameter are selected based on their performance on the validation dataset. This intermediate step reduces the total search space of the parameters and allows for better interpretation and analysis of the effect of the other parameters on the performance of the developed models. The testing set is used to select the granularity-metric and sub-bands combinations. 

All the parameters were selected on the Leak\_process dataset based on their performance on its respective validation and testing sets. The best performing model, encompassing the \{granularity, sliding window, metric\} parameters and SVC hyper-parameters (regularization parameter, kernel, and gamma), is evaluated on the datasets with different ambient conditions, denoted by Leak\_noprocess and NoLeak\_noprocess datasets. 

\begin{table}[htbp]
\centering
\begin{tabular}{|c|c|}
\hline
\textbf{Parameter} & \textbf{Values}           \\[0.5ex] \hline 
granularity         & \{1000, 2000, 5000\}      \\[0.5ex] \hline
sliding window      & \{3, 5, 7\}               \\[0.5ex] \hline
SVC\_kernel         & \{linear, rbf\}           \\[0.5ex] \hline
SVC\_Cost             & \{1, 10, 100, 1000\}      \\[0.5ex] \hline
SVC\_gamma          & \{1, 0.1, 0.001, 0.0001\} \\[0.5ex] \hline
metrics             & \{mean, median, iqr\}     \\[0.5ex] \hline
\end{tabular}
\caption{Experimental Parameters}
\label{Experimental Parameters}
\end{table}

\section{Results}
\begin{figure*}[htbp]
    \centering
    \begin{subfigure}{0.48\textwidth}
        \centering
        \includegraphics[scale=0.17]{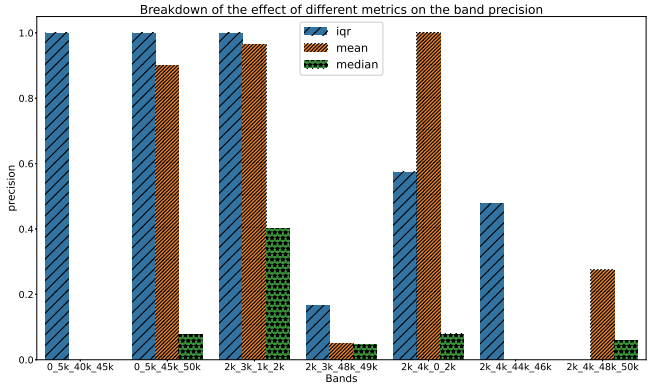}
        \caption{Precision on the testing set}
        \label{testing_precision}
    \end{subfigure}
    \hfill
    \begin{subfigure}{0.48\textwidth}
        \centering
        \includegraphics[scale = 0.16]{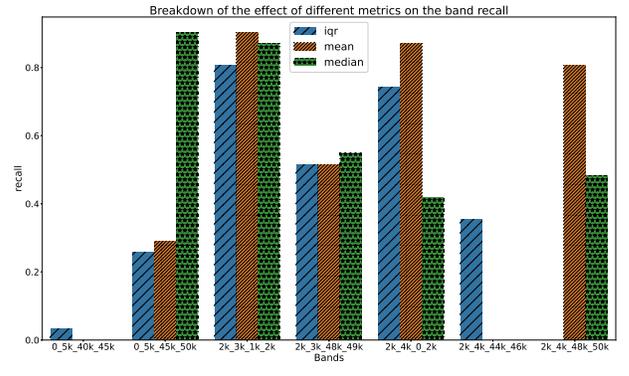}
        \caption{Recall on the testing set}
        \label{testing_recall}
    \end{subfigure}
    \caption{Effect of metrics on the performance of sub-bands on the testing set of the Leak\_process dataset}
    \label{testing_performance_p}
\end{figure*}
This section first outlines the parameters that best performed on the dataset with industrial noise. After that, the results of applying the obtained models to datasets with no industrial noise are evaluated. 

\subsection{Parameter Selection and Hyper-parameter Optimization}
The best overlapping time window and SVC hyper-parameters are obtained based on their performance on the validation set.  The selected hyper-parameters are the linear kernel and the 1 penalizing cost and 7 as the overlapping time window. The validation step has disqualified some granularity-metric and band combinations, but it still did not provide a clear-cut favorite set of combinations. As such, several combinations qualified for the next step, which were evaluated on the testing set of the Leak\_process dataset. 

Figure \ref{testing_performance_p} depicts the results of applying these combinations to the testing set. Due to space restrictions, the precision and recall were only taken as metrics for evaluating the models. The discrepancy in performance for different metrics of the same band shows the effect of outliers on the performance of different band combinations. In most cases, the mean metric displayed the better results compared with outlier-resistant metrics such as median and iqr. With regard to band combinations, the combinations of low and high bands, such as band\_0\_5k\_40k\_45k, have displayed poor recall results when compared to combinations involving lower bands. Given that these band combinations are selected based on their correlation with leaks, the obtained results show that high bands have no contribution to identifying leaks. Instead, these high correlation values are attributed to the effect of industrial noise. These results show the importance of the inclusion of multiple bands for identifying leaks and the effect of industrial noise in emulating them. 

The only two combinations that have displayed high precision and recall are the combinations of iqr and band\_2k\_3k\_1k\_2k and mean and band\_2k\_4k\_0\_2k. These combinations qualify for the subsequent analysis stage that takes the models built on them and applies them to the Leak\_noprocess and NoLeak\_noprocess datasets.
\subsection{Best Performing Models Implementation}
The two best performing combinations are trained on the whole Leak\_process dataset and evaluated on the Leak\_noprocess and NoLeak\_noprocess datasets. This approach assesses the generalizability of the model developed to be applied to conditions with no industrial noise as it has proven its merit in its presence. The parametric results of the given model are provided in Table  \ref{best_model_iqr} and \ref{best_model_mean}. The results show that both models managed to identify most of the leaks, reflected by high recall, while avoiding false alarms, reflected by high precision. Applied to the Leak\_noprocess dataset, the two models successfully identified all the normal cases. For the Leak\_process dataset, slight differences in results between the models can be highlighted. The results show that the combination of mean and band\_0\_2k\_2k\_4k outperforms the other combination in identifying leaks, which demonstrate that the wider bands are more informative in terms of leak detection. Additionally, this result shows that leaks can be better detected when incorporating outlier calculation. This conclusion is emphasized by the better results displayed by the combination that includes the mean metric. 

\begin{table}[htbp]
\centering
\scriptsize
\begin{tabular}{|c|c|c|}
\hline
\textbf{Metrics} & \textbf{Leak\_noprocess} & \textbf{NoLeak\_noprocess} \\ \hline
accuracy & 0.99 & 1 \\[0.5ex] \hline
recall & 0.82 & N/A \\[0.5ex] \hline
precision & 1 & N/A \\[0.5ex] \hline
specificity & 1 & 1 \\[0.5ex] \hline
\end{tabular}
\caption{Results for iqr and band\_1k\_2k\_2k\_3k}
\label{best_model_iqr}
\end{table}

\begin{table}[htbp]
\centering
\scriptsize
\begin{tabular}{|c|c|c|}
\hline
\textbf{Metrics} & \textbf{Leak\_noprocess} & \textbf{NoLeak\_noprocess} \\ \hline
accuracy & 0.99 & 1 \\[0.5ex] \hline
recall & 0.88 & N/A \\[0.5ex] \hline
precision & 1 & N/A \\[0.5ex] \hline
specificity & 1 & 1 \\[0.5ex] \hline
\end{tabular}
\caption{Results for mean and band\_0\_2k\_2k\_4k}
\label{best_model_mean}
\end{table}
\section{Conclusion}
This paper addressed detecting leakages in fluid-carrying pipes in a harsh industrial environment. This environment is distinct compared to other urban and isolated environments in terms of the inclusion of interfering noise and the deployment requirements of any detection module. The former condition challenges the identification of leaks. The latter results in data scarcity imposing the use of low-complexity ML techniques. To address these limiting conditions, this work proposes an ML pipeline encompassing feature selection techniques and time-based feature engineering. The resulting data is fed to an SVC algorithm that satisfies the environment's deployment requirements. The models obtained following the proposed ML pipeline successfully identified leaks in environments with and without industrial noise. While the results obtained are promising, there is still considerable room for improvement. A suggested method would evaluate the proposed approach using various light-weight ML classification models such as Decision Trees and ensemble learners to compare them to the currently developed SVC model. Additionally, it is critical to simulate leaks and industrial noise to increase the amount of available data, allowing for a more extensive and profound evaluation. 

\bibliographystyle{IEEEtranN}
\bibliography{refs}
\vspace{12pt}

\end{document}